\documentclass[conference]{IEEEtran}
\IEEEoverridecommandlockouts
\usepackage{cite}
\usepackage{amsmath,amssymb,amsfonts}
\usepackage{algorithmic}
\usepackage{graphicx}
\usepackage{textcomp}
\usepackage{xcolor}
\pagecolor{white}
\usepackage{bbold}
\usepackage{multirow}
\usepackage{bm}
\def\BibTeX{{\rm B\kern-.05em{\sc i\kern-.025em b}\kern-.08em
    T\kern-.1667em\lower.7ex\hbox{E}\kern-.125emX}}
\begin{document}

\title{Multilingual Transformer Encoders: a Word-Level Task-Agnostic Evaluation}

\author{\IEEEauthorblockN{Félix Gaschi}
\IEEEauthorblockA{\textit{team ORPAILLEUR} \\
\textit{LORIA}\\
Vandœuvre-lès-Nancy, France \\
felix.gaschi@loria.fr}
\and
\IEEEauthorblockN{François Plesse}
\IEEEauthorblockA{\textit{R\&D dpt.} \\
\textit{Posos}\\
Paris, France \\
francois@posos.fr}
\and
\IEEEauthorblockN{Parisa Rastin}
\IEEEauthorblockA{\textit{team ABC} \\
\textit{LORIA}\\
Vandœuvre-lès-Nancy, France \\
parisa.rastin@loria.fr}
\and
\IEEEauthorblockN{Yannick Toussaint}
\IEEEauthorblockA{\textit{team ORPAILLEUR} \\
\textit{LORIA}\\
Vandœuvre-lès-Nancy, France \\
yannick.toussaint@loria.fr}
}

\IEEEoverridecommandlockouts
\IEEEpubid{\makebox[\columnwidth]{978-1-7281-8671-9/22/\$31.00~\copyright2022 IEEE \hfill}
\hspace{\columnsep}\makebox[\columnwidth]{ }}

\maketitle

\begin{abstract}
     Some Transformer-based models can perform cross-lingual transfer learning:
    those models can be trained on a specific task in one language and give
    relatively good results on the same task in another language, despite having
    been pre-trained on monolingual tasks only. But, there is no consensus yet on
    whether those transformer-based models learn universal patterns across
    languages. We propose a word-level task-agnostic method to evaluate the
    alignment of contextualized representations built by such models. We show
    that our method provides more accurate translated word pairs than previous
    methods to evaluate word-level alignment. And our results show that some
    inner layers of multilingual Transformer-based models outperform other
    explicitly aligned representations, and even more so according to a
    stricter definition of multilingual alignment. \looseness=-1

\end{abstract}


\section{Introduction}

Building aligned multilingual representations is namely useful for cross-lingual document retrieval. 
For example, a query written in German might be answered with documents in German but also in other languages, like scientific papers in English. Aligned multilingual 
representations would allow to embed all documents as well as the query in a shared space, allowing document retrieval by nearest-neighbor search. Good word-level alignment could lead to more fine-grained 
cross-lingual information retrieval, like passage retrieval, allowing for extracting relevant passages from 
relevant documents in multiple languages. \looseness=-1

This paper provides an evaluation of the word-level
multilingual alignment of contextualized representations produced by
multilingual models, particularly by multilingual Transformer encoders \cite{vaswani-etal-20173}. This
evaluation framework does not rely on any specific probing task and directly compares word representations instead of sentence-level ones \cite{pires-etal-2019-multilingual,singh-etal-2019-bert}. The proposed method relies on a bilingual dictionary instead of a probabilistic tool like FastAlign \cite{dyer-etal-2013-simple}, as in \cite{zhao-etal-2021-inducing}, to retrieve more accurate word pairs in parallel texts. It
aims at measuring directly how well word representations from different
languages are aligned and at producing a comparative analysis of existing models.

To build accurate language representations, models based on the Transformer architecture were
introduced a few years ago with BERT \cite{devlin-etal-2019-bert} and now play a
key role in natural language processing (NLP) research. Those models are first
pre-trained in an unsupervised manner on large text datasets and can then be
fine-tuned on a downstream task like sentence classification, named entity
recognition (NER), or extractive question answering. BERT is also used as a building block for document retrieval models like ColBERT \cite{colbert}.

NLP research is mainly led around the English language \cite{Bender_2011}. BERT,
and many of its proposed variations like RoBERTa
\cite{DBLP:journals/corr/abs-1907-11692}, SciBERT \cite{Beltagy2019SciBERT}, or
BioBERT \cite{DBLP:journals/corr/abs-1901-08746}, are pre-trained and evaluated
solely in English. Some BERT-like models were proposed in other languages, like
CamemBERT \cite{martin-etal-2020-camembert} and FlauBERT \cite{le-etal-2020-flaubert-unsupervised} in French, FinBERT \cite{ronnqvist-etal-2019-multilingual} (Finnish), AfriBERT \cite{ralethe-2020-adaptation} (Afrikaans), etc.
They bring the advances of Transformer-based models to yet another language but they are still monolingual.

Multilingual Transformer-based models, like mBERT \cite{devlin-etal-2019-bert}
or XLM-R \cite{conneau-etal-2020-unsupervised}, are the focus of our paper.
They are pre-trained with the same objective as their monolingual
counterparts but on several monolingual corpora in different languages.

Although they have never been explicitly trained on parallel text, those models
were shown to be able to generalize well from one language to another. For instance, instead of fine-tuning and evaluating in the same language after
pre-training, the models can be fine-tuned on an English Named Entity Recognition
(NER) training dataset and evaluated on a French one and still obtain good
results. This process was dubbed "Cross-lingual Transfer Learning"
(CTL)\cite{pires-etal-2019-multilingual}.

Despite those cross-lingual generalization abilities, there is no consensus on
whether those models exhibit aligned multilingual representations. By aligned
representations, we mean that a word and its translation should be attributed similar representations, in the same way that
words or sentences of similar meaning should be represented by similar
representations in a monolingual embedding.

We want to directly evaluate the quality of the alignment in multilingual models
because cross-lingual transfer learning is not necessarily correlated with
multilingual alignment. Our contribution is (1) to propose a method for extracting pairs of translated words
in context; (2) to show that the proposed method extracts more accurate pairs than others \cite{zhao-etal-2021-inducing} which rely on word-alignment tools like FastAlign\cite{dyer-etal-2013-simple}; and (3) to reveal that the alignment produced by most Transformer-based
multilingual models is competitive with other representations.

\section{Related Work}

Building universal multilingual representations has been envisioned long before
the introduction of the Transformer architecture with ideas like Chomsky's
``Universal Grammar" \cite{chomsky_1968} and the ``linguistic universals" of
Greenberg \cite{greenberg1966language}. But Transformer-based models are
expected to build such representations ``at scale", according to 
\cite{conneau-etal-2020-unsupervised}, with the goal that low-resource languages
might benefit from high-resource ones by sharing some features like common
sub-words \cite{pires-etal-2019-multilingual}. However, many works have focused on the ability to perform
cross-lingual transfer learning
\cite{pires-etal-2019-multilingual,wu-dredze-2019-beto,conneau-etal-2020-unsupervised}
and less on the multilingual alignment of the produced representation \cite{DBLP:journals/corr/abs-2107-00676}.

\subsection{Word Embedding Alignment}

When word embeddings were introduced \cite{mikolov2013efficient}, it was then
shown that two or more monolingual embeddings could be aligned effectively
\cite{DBLP:journals/corr/MikolovLS13}. State-of-the-art alignment methods for
word embeddings give impressive results on bilingual lexicon induction (BLI), a
task where the translation of a word is retrieved with nearest-neighbor search
\cite{muse_conneau2017word,joulin-etal-2018-loss,artetxe-etal-2018-robust}. In
our experiments, a specific version of those aligned word
embeddings will be used as a baseline: monolingual FastText embeddings \cite{bojanowski2017enriching} aligned with
RCSLS \cite{joulin-etal-2018-loss}. 

A multilingual word embedding provides a static representation of each word,
meaning that a word will always be represented by the same vector regardless of 
the context, whereas the Transformer-based models which are the focus of our paper provide a contextualized representation of a word. The representation attributed to a word by such a deeper model varies according to the surrounding words.

\subsection{Multilingual Language Models}

Deeper multilingual language models were first proposed as a way to perform
unsupervised machine translation \cite{lample-etal-2018-phrase}. Besides
unsupervised translation models, LASER \cite{DBLP:journals/corr/abs-1812-10464} is a multilingual sentence embedding based on
RNNs and trained on several parallel corpora. MultiFiT
\cite{eisenschlos-etal-2019-multifit} is a training procedure involving LASER and a monolingual model
to perform a cross-lingual transfer. However, we
will focus here on Transformer-based models which are pre-trained with fewer parallel texts and are not built specifically for sentence-level downstream tasks like LASER and MultiFit.

\subsection{Multilingual Transformer Encoders}

The Transformer architecture was initially proposed as an encoder-decoder
architecture \cite{vaswani-etal-20173}, for sequence-to-sequence tasks like translation. But
BERT \cite{devlin-etal-2019-bert} uses only the encoder part of the architecture
to set a new standard for tasks that require encoding the full sentence like
sentence classification, named entity recognition, or extractive question
answering. BERT is typically pre-trained to predict randomly masked-out words
across a wide English corpus. A multilingual version of BERT, called mBERT, was
proposed and is the same architecture pre-trained for the same task but on
104 monolingual corpora of distinct languages.

Despite not being pre-trained explicitly on parallel text, mBERT was shown to
 have surprisingly good cross-lingual transfer abilities
 \cite{pires-etal-2019-multilingual, wu-dredze-2019-beto}. For example, when
 mBERT is fine-tuned on a classification task in English, it gives competitive
 results when evaluated in French. This has also been shown to work between
 typologically different languages, although to a lesser extent than similar
 languages.

Multilingual language models (MLLMs) are often either framed as a way to perform
unsupervised machine translation or either as a ground for generalizing the
learning of downstream tasks across languages. In this paper, we wonder whether MLLMs can hold ``universal" patterns
across languages, or at least patterns that are common to a wide range
of different languages, and produce aligned representations.

\subsection{Absence of consensus about multilingual alignment}

Despite the surprisingly efficient cross-lingual transfer of models like mBERT,
there is no consensus on whether those multilingual models learn universal
multilingual patterns, as a recent literature review states \cite{DBLP:journals/corr/abs-2107-00676}. With some
probing tasks, it is shown that we can learn syntactic trees, which represent the grammatical structure of a sentence, that are consistent
across languages with heuristics applied to attention activation values
\cite{limisiewicz-etal-2020-universal}. But other probing tasks show that
the same models retain language-specific information at every layer
\cite{wu-dredze-2019-beto,DBLP:journals/corr/abs-2009-12862}.

\begin{figure}
    \centering
    \includegraphics[width=\columnwidth]{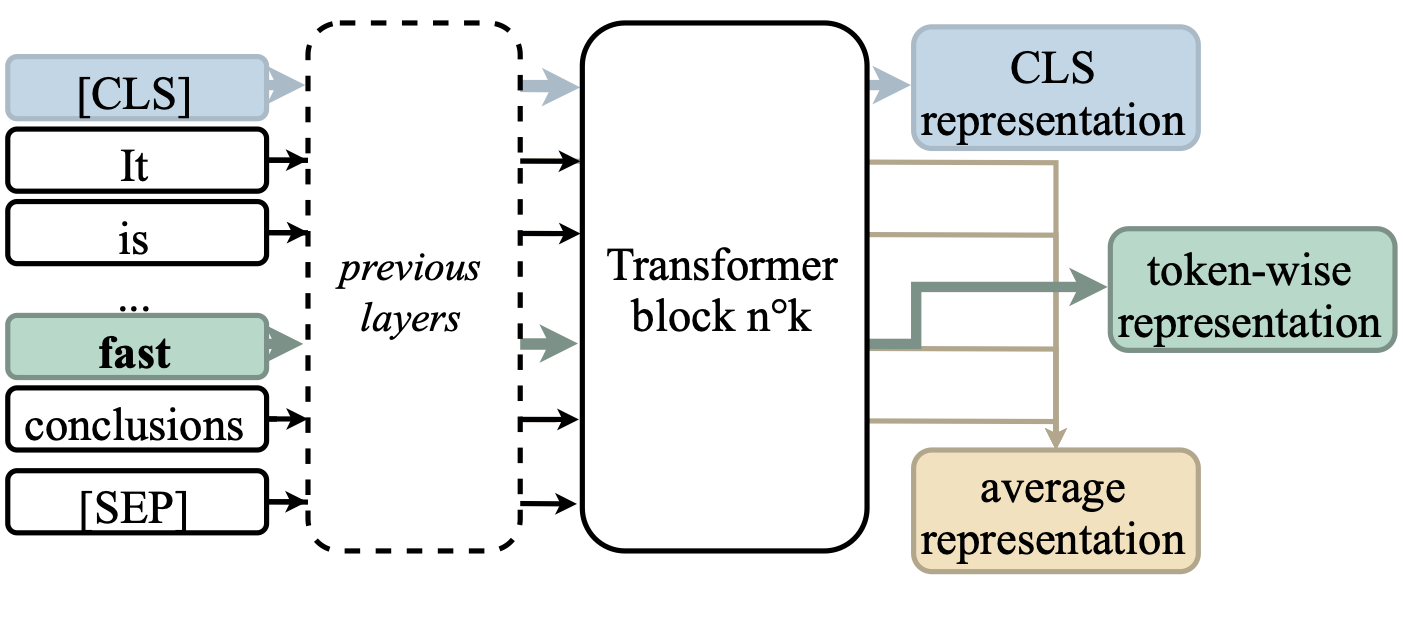}
    \caption{Different representations extracted from multilingual Transformers:
    sentences representations
    \cite{singh-etal-2019-bert,pires-etal-2019-multilingual} or token
    representations, ours and
    \cite{zhao-etal-2021-inducing}.\label{fig:representations}}
\end{figure}

Another approach for studying multilingual patterns learned by a model
is to perform inferences on parallel text. Different
approaches lead to opposite results
\cite{pires-etal-2019-multilingual,singh-etal-2019-bert}. As shown in Fig.
\ref{fig:representations}, \cite{pires-etal-2019-multilingual} compares sentence
representations obtained by averaging sub-word representations, whereas
\cite{singh-etal-2019-bert} uses the representation of the initial token [CLS]
which is supposed to hold a representation of the sentence. The former found
aligned representations, while the latter concluded that mBERT ``is not an
interlangua". We propose to compare word representations instead similarly to \cite{zhao-etal-2021-inducing}.

Indeed, even if the sentence-level alignment was guaranteed, does it necessarily
mean that there is a word-level multilingual alignment in the representations
produced by mBERT and others? the CLS representation is independent of the token representation. And if the same information was repeated in each token embedding, one could have an alignment of average sentence representation, but not of the single tokens.

It was shown that mBERT builds representations that enable the extraction of
corresponding words in a pair of translated sentences with fair accuracy, and
AWESOME \cite{DBLP:journals/corr/abs-2101-08231} was proposed to fine-tune mBERT
and improve this word-level matching. In another work
\cite{zhao-etal-2021-inducing}, translated pairs of words were extracted from
translated sentences with FastAlign \cite{dyer-etal-2013-simple}. the similarity
between the contextualized representations built by mBERT for those words was
compared with the similarity between representations of random words. They
observed that the similarity for translated words was close to that of random
pairs. We show that tools like FastAlign make too many alignment mistakes to be
able to conclude about the quality of the multilingual alignment produced by models
like mBERT. Our method relies on the extraction of translated words from
translated sentences in a manner that is less prone to errors for evaluating
multilingual models.

\section{Methodology}

\label{sec:methodology}

\subsection{Translated-in-context word pairs}

In order to extract translated words from translated sentences with a minimized number of errors,
we start from a bilingual dictionary. And instead of trying to extract all possible word pairs 
with a word-alignment tool like FastAlign, we extract only those that we are certain of.

\begin{figure}
    \centering
    \includegraphics[width=0.8\columnwidth]{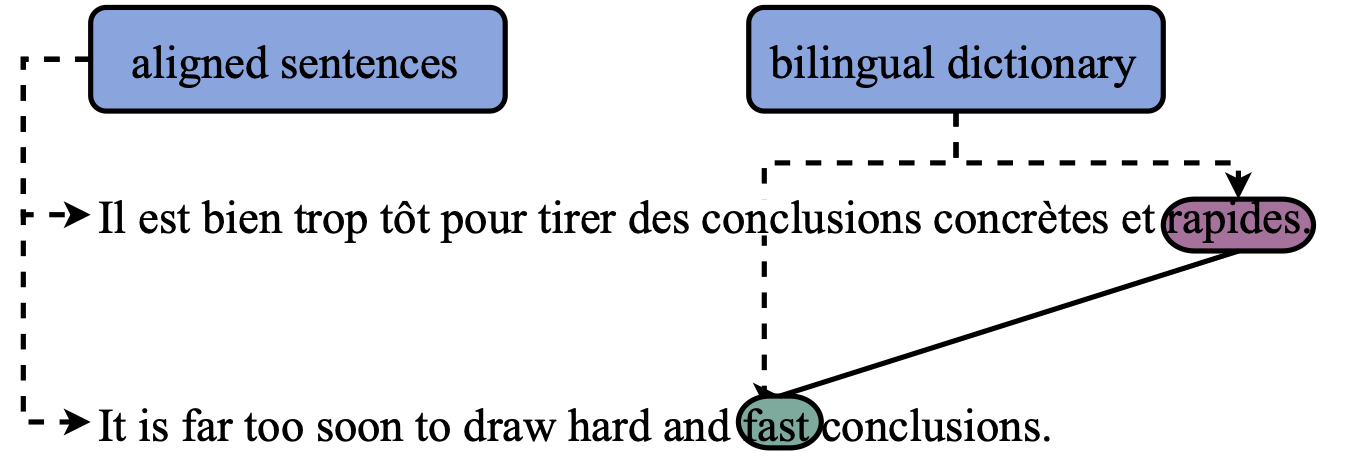}
    \caption{Extracting a contextualized translated pair with a bilingual dictionary.\label{fig:method}}
\end{figure}

Our method requires two datasets: a translation dataset containing gold standard translated
pairs of sentences and a bilingual dictionary. The bilingual dictionary contains pairs of
translated words, like the pair "rapide"-"fast" in our example Fig.
\ref{fig:method}. In a pair of translated sentences from the translation
dataset, for each word from one sentence, every potential translated word
indicated by the bilingual dictionary is collected from the other sentence. A
pair of translated words is kept with its context if there
is only one candidate for the translation.

The pairs obtained can be seen as a contextualized bilingual dictionary, where
translated words go along with their context. We argue that we could not use
pairs of words from the original bilingual dictionary directly because the
models we want to evaluate are pre-trained on whole sentences and, since they
have not been pre-trained on single words, they might not give relevant
representations of words without context.

Working at the word level instead of the sentence level like
\cite{singh-etal-2019-bert} and \cite{pires-etal-2019-multilingual} is chosen
for several reasons. A good alignment of sentence representations does
not necessarily guarantee a good alignment of word representations. Also, 
working at the word level allows a more direct comparison with multilingual word
embeddings, which can then be used as a baseline.

\subsection{Inference on sentences}

Once we have extracted translated-in-context pairs of words, they are passed
through the model we want to evaluate to produce contextualized representations of the translated words.

To build the representation of each contextualized word, the whole sentences
are passed to the evaluated model (typically mBERT) and the contextualized representations of the words from the
extracted word pairs are kept. We can extract such representations for
each stacked Transformer block (or layer). By convention, the representation of index 0
will be the one from the input un-contextualized embedding layer. For example,
mBERT, which uses 12 Transformer blocks, will produce 13 representations for a
word, the 0th one for the initial embedding and the 1st to 12th for the output
of each Transformer block.

\subsection{Nearest-neighbor retrieval task}

For each layer of a Transformer-based model, a nearest-neighbor retrieval
task is then performed on the produced representations. In a similar manner as
\cite{pires-etal-2019-multilingual}, a fixed number $N$ of representations of
translated-in-context pairs is randomly sampled. Then, for each of those sampled
pairs of representations $(u_i, v_i)$, the aim is to check that $v_i$ is the
nearest neighbor of $u_i$ with respect to all other $v_j$. The final evaluation
score is then the proportion of pairs for which the translation is the nearest
neighbor. More formally our score is given by:

\begin{equation}
    S_{\text{weak}}(U,V) = \frac{1}{N} \sum_{i = 0}^N \mathbb{1}\left[ s(u_i, v_i) > \max_{j\neq i} s(u_i, v_j) \right]\label{eq:weak}
\end{equation}

Where $U$ and $V$ are the sets of sampled $u_i$ and $v_i$ respectively, $s$ is a
similarity function or retrieval criterion and $N$ the number of sampled pairs.

The retrieval criterion used is the cross-domain similarity local scaling (CSLS)
\cite{muse_conneau2017word}. It is a modified cosine similarity commonly used
for retrieval in word embeddings \cite{muse_conneau2017word,joulin-etal-2018-loss,artetxe-etal-2018-robust} which takes into account the density around the
compared representations by deducing from the initial similarity the averaged
cosine similarity of the $k$ nearest neighbor of each representation in the
pair: 

\begin{equation}
    \begin{split}
    s(u_i, v_j) = 2 \cos(u_i, v_j) - \frac{1}{k} \sum_{v \in \mathcal{N}_V^k (u_i)} \cos(u_i, v)\\ - \frac{1}{k} \sum_{u \in \mathcal{N}_U^k (v_j)} \cos(u, v_j)
    \end{split}\label{eq:csls}
\end{equation}

$c$ is the cosine similarity and $\mathcal{N}_V^k(u_i)$ the set of $k$ nearest
neighbors of $u_i$ in $V$.

In our reported experiments, we used a similarity based on the cosine similarity
and not on the l2-distance for two reasons: (1) the CSLS criterion is commonly
used for evaluating multilingual word embeddings which is our baseline and (2)
we noticed that l2-distance gives similar results for the evaluated models which
can be explained by the Transformer architecture using a layer normalization at
the end of each Transformer block. If the representations have roughly
constant norms, then l2-distance is approximately proportional to cosine
distance.

\subsection{Evaluating strong alignment}

\begin{figure}
    \centering
    \fbox{\includegraphics[width=0.8\columnwidth]{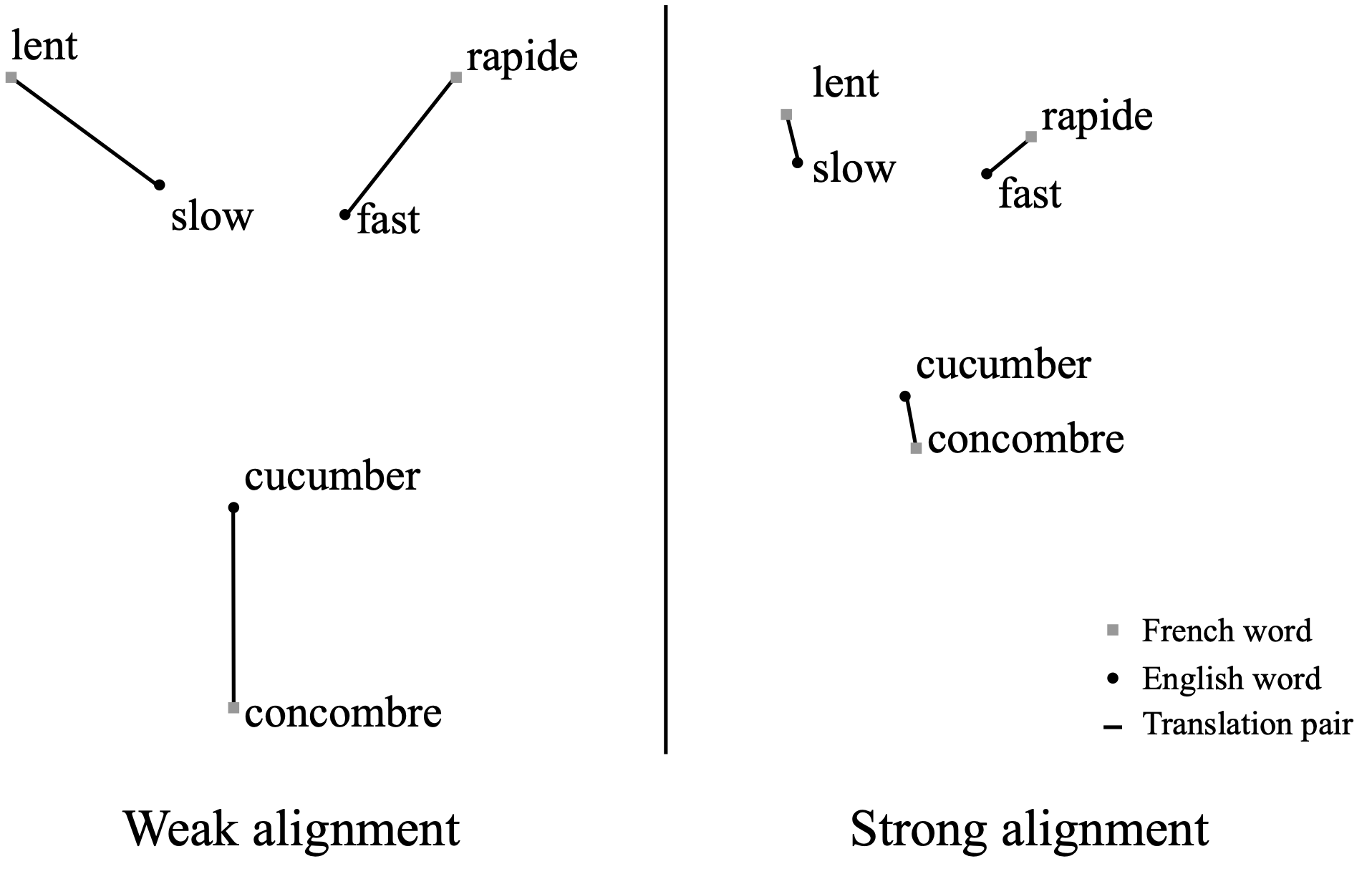}}
    \caption{Toy example illustrating weak and strong alignment.}
    \label{fig:strong_vs_weak}
\end{figure}

With the retrieval score $S_{\text{weak}}$ described in Equation \ref{eq:weak},
we are only evaluating "weak" alignment as described by
\cite{roy-etal-2020-lareqa}. Weak alignment is defined as the fact that the
nearest neighbor in language $L_1$ of any item in a language $L_2$ is the most
relevant item (i.e. the translation in our case). For example, as shown in Fig. \ref{fig:strong_vs_weak}, "fast" will be
closer to "rapide" than "slow" or "cucumber" but "lent" or
"concombre" can be even closer.

Strong alignment is defined as the fact that translated pairs are closer than
irrelevant pairs, regardless of the language. In this case, "fast" should be
closer to "rapide" than "slow" and "cucumber", but also than "lent" and
"concombre". Strong alignment is useful for document retrieval. Indeed, if a
document retrieval system is expected to answer a query with documents from
different languages, it should be able to rank documents across languages. 

We propose another retrieval score like in Eq. \ref{eq:weak} which
evaluates strong alignment by checking that for each pair of representations
$(u_i, v_i)$ the translation $v_i$ is closer to $u_i$ than any element of $U$
instead of $V$:

\begin{equation}
    S_{\text{strong}}(U,V) = \frac{1}{N} \sum_{i = 0}^N \mathbb{1}\left[ s(u_i, v_i) > \max_{j\neq i} s(u_i, {\bm u_j}) \right]\label{eq:strong}
\end{equation}

The only change is in bold: $v_j$ has been replaced by $u_j$. For instance, we want the representation of the French word "rapide" to be closer to its translation "fast" than any other English word (instead of French for $S_{\text{weak}}$).

\section{Experimental Setup}
\label{sec:models}
We compare six models:

\paragraph{mBERT} \cite{devlin-etal-2019-bert} pre-trained on Wikipedia in the 104 most frequent languages with two objectives: (1) Masked language modeling (MLM); predicting
randomly masked out words and (2) next sentence prediction (NSP) determining
whether two sentences are consecutive only using the representation of the [CLS]
token (Fig. \ref{fig:representations}). It was not trained on any parallel text.

\paragraph{XLM-R} \cite{conneau-etal-2020-unsupervised} pre-trained on
CommonCrawl (which contains Wikipedia) in 100 languages. It is based on RoBERTa
\cite{DBLP:journals/corr/abs-1907-11692}, so it only pre-trains with the MLM
objective. It was not trained on any parallel text.

\paragraph{XLM-15 (MLM+TLM)} \cite{DBLP:journals/corr/abs-1901-07291} pre-trained on
Wikipedia in the 15 languages from the XNLI dataset
\cite{DBLP:journals/corr/abs-1809-05053}. It is pre-trained on MLM objective but
also on parallel text (drawn from XNLI) using the Translated Language Modeling
(TLM) objective, which is an MLM objective applied to parallel text, to allow the
model to attend to words from the other language to predict masked out words.
Additionally to its training on parallel data, its input embedding is added to a
language embedding indicating the language of the sentence, whereas models like
mBERT and XLM-R have no input information about the language.

\paragraph{XLM-100} \cite{DBLP:journals/corr/abs-1901-07291} pre-trained on
Wikipedia in 100 languages with MLM only.

\paragraph{AWESOME} \cite{DBLP:journals/corr/abs-2101-08231} which is mBERT
fine-tuned on a variety of self-supervised objectives and supervised objectives
on a parallel corpus to improve word-level alignment for extracting pairs of translated words
in parallel sentences: MLM, TLM but also objectives on the consistency of the produced alignment.

\paragraph{mBART} \cite{DBLP:journals/corr/abs-2001-08210} contrary to all
previously mentioned models which are Transformer encoders, mBART follows an
encoder-decoder architecture. It was pre-trained on
filling missing spans of texts for 50 languages. We consider only the representations built by the encoder part of the model as we empirically observed that 
the decoder gives worse multilingual alignment than the encoder.


To evaluate the multilingual alignment of those models, we rely on the WMT19 dataset
\cite{wmt19translate} for parallel sentences. And For the
bilingual dictionaries, we use MUSE \cite{muse_conneau2017word}. Monolingual FastText embeddings \cite{bojanowski2017enriching} aligned with
RCSLS \cite{joulin-etal-2018-loss} are used as a baseline.

For all experiments, the number of sampled pairs is $N=5 000$ as in \cite{pires-etal-2019-multilingual}. The number of neighbors for the CSLS criterion (cf. Eq. \ref{eq:csls}) is $k=10$ as in \cite{joulin-etal-2018-loss}. To avoid favoring contextualized models over FastText aligned embedding we chose to sample distinct pairs of words. We empirically verified for all the layers of three models (mBERT, XLM-R, AWESOME) on three language pairs and for 10 different sampling of pairs of words that it gives equivalent results: we observed a strong correlation with a 0.86 Spearman rank correlation (p-value $<0.01$). To obtain the 95\% confidence intervals on all figures and the empirical standard deviation for all tables, we perform 10 runs of each experiment.


\section{Results}

In this section, we first investigate the fact that different results were
obtained regarding sentence-level alignment. Then, we demonstrate that our
method provides better pair of words than FastAlign thanks to a lesser number of carefully selected pairs.
Finally, we perform the word-level evaluation allowed by our method for weak
alignment first and then for strong alignment, showing that multilingual
Transformer-based models bring better alignment than multilingual word
embeddings according to our metrics. 

\subsection{The chosen sentence representation influences the results}

Before reporting our results on word-level alignment, we investigate the
contradiction between \cite{pires-etal-2019-multilingual} and
\cite{singh-etal-2019-bert} on sentence-level alignment for the mBERT model. 

On the one hand, \cite{pires-etal-2019-multilingual} performed a nearest-neighbor
search similar to ours on sentence representations. Each sentence was
represented by the average of the embeddings of its tokens and those vectors
were centered for each language. High retrieval accuracy is observed with this
method for typologically similar languages. \looseness=-1

On the other hand, \cite{singh-etal-2019-bert} reported a Canonical Correlation
Analysis (CCA) across layers of the representation in various contexts of the
initial token [CLS] which is expected to encode the meaning of the sentence.
This method shows that those representations are dissimilar, and the
dissimilarity grows stronger towards the deeper layers.

\begin{figure}[!t]
    \centering
    \includegraphics[width=0.8\columnwidth]{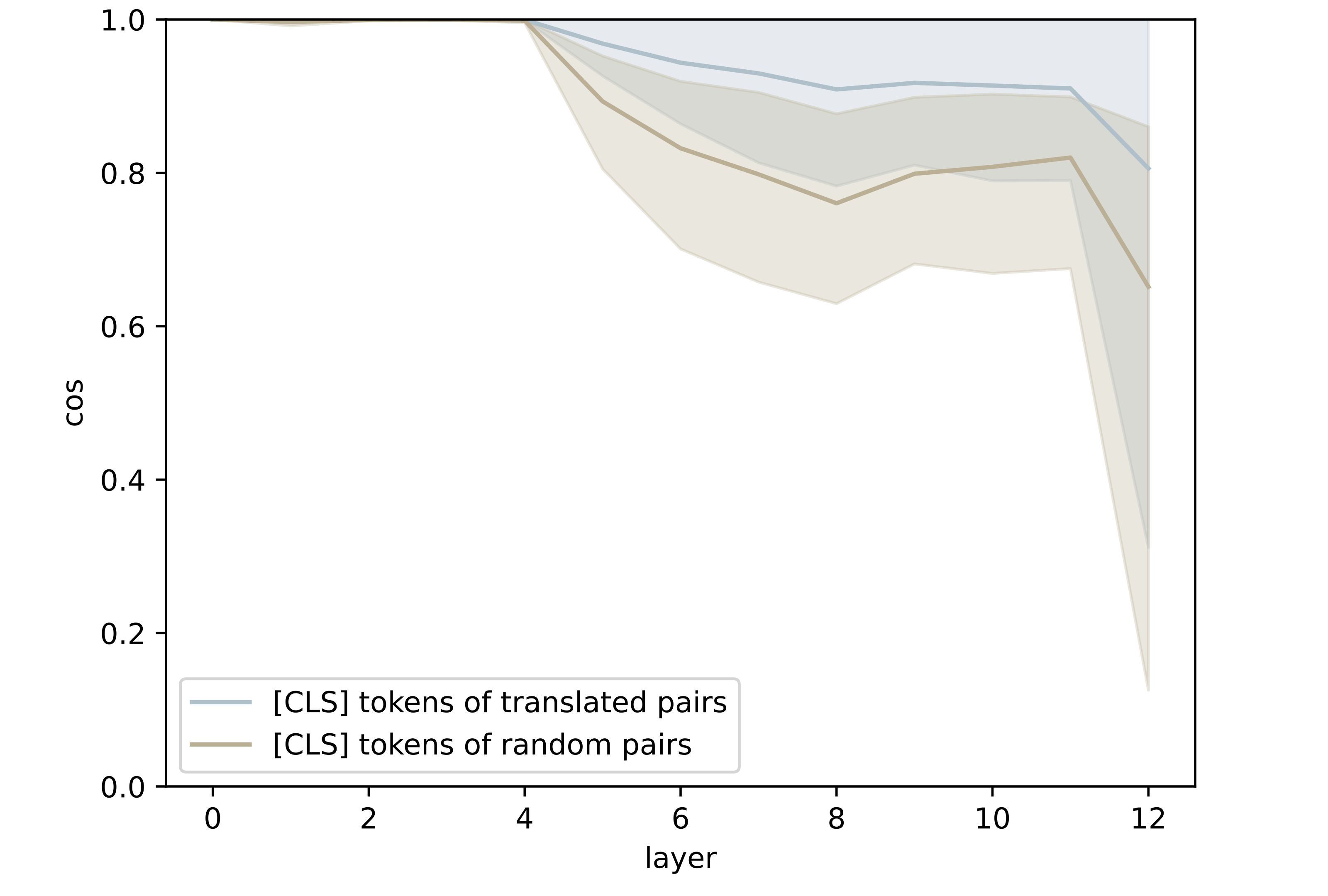}
    \caption{Evolution of the similarity between CLS tokens across layers of mBERT for translated and random pairs (ru-en), with 95\% confidence interval.\label{fig:cls_sim}}
\end{figure}

On Fig. \ref{fig:cls_sim} we observe the same decrease across layers, as \cite{singh-etal-2019-bert}, for the similarity between [CLS] tokens
of translated sentences  with the
cosine similarity instead of the CCA. But the similarity decreases even more for an equal number
of random pairs of sentences drawn from the same dataset. 

The decrease of the similarity between translated pairs can be deduced from the
fact that it reaches exactly 1 at the 0th layer, corresponding to
non-contextualized embeddings which will have the same value if the token is the
same, here "[CLS]". The similarity can do nothing but decrease when information
from the context is injected into the contextualized representation of the [CLS] token.

However, to fairly compare representations based on the CLS token and averaged
representations of the sub-words, they can both be used in the same nearest-neighbor
task as proposed in our method for word representations. We apply Equation \ref{eq:weak} to sentence
representations. The sentence representations are not centered as in
\cite{pires-etal-2019-multilingual} as we want to evaluate directly the quality of
alignment, and not to artificially improve it.

\begin{table}[!t]
    \centering
    \caption{NN-search for sentence representations\label{tab:sent_nn}}
    \begin{tabular}{l|ccc}
  \hline
method and layer & de-en & ru-en & zh-en\\
\hline
FastText avg & 53.3 (1.1) & 26.1 (1.1) & 1.4 (0.2)\\
\hline
CLS first & 0.0 (0.0) & 0.0 (0.0) & 0.0 (0.0)\\
avg first & 56.1 (2.1) & 17.8 (0.7) & 41.7 (1.1)\\
\hline
CLS best & 77.9 (0.4) & 59.6 (0.5) & 51.1 (0.5)\\
avg best & \textbf{90.1} (0.2) & \textbf{82.1} (0.4) & \textbf{88.1} (0.6)\\
\hline
CLS last & 60.9 (0.8) & 40.5 (0.5) & 21.0 (0.8)\\
avg last & 87.3 (0.4) & 75.5 (0.5) & 79.4 (0.3)\\
\hline

    \end{tabular}
\end{table}

Results reported in Table \ref{tab:sent_nn} show that whatever the
chosen layer is, averaged representations of the sentences provide better
multilingual alignment than CLS representations. And for deeper layers, both give
better alignment than aligned word embeddings averaged over the sentence. It must also 
be noted that alignment of the averaged mBERT representation suffers less from the typological distance between 
languages than the CLS representation or aligned FastText.

Sentence representations can give different results according to the chosen
method, but it seems that whatever it is, sentence representations produced by
mBERT are relatively aligned across languages. However, the CLS representation
seems to be less relevant. Furthermore, the CLS token does not exist in all
Transformer-based models, hence we recommend relying on averaged representations
instead.

\subsection{Using bilingual dictionary over alignment tools}

Fair multilingual alignment of sentence representations does not guarantee a
good word-level multilingual alignment. In Section \ref{sec:methodology}, we argued that the proposed method is
a way to extract translated words which is less prone to errors. Table
\ref{tab:precision} shows the proportion of accurate pairs extracted by our
method and FastAlign \cite{dyer-etal-2013-simple}. 
It demonstrates that our method extracts proportionally more accurate pairs than
FastAlign, although it provides fewer pairs in quantity to be fair. For 10 000 
sentences from WMT19 for the English-German pair: our method extracts 50 590
word pairs and 190 665 for FastAlign.

\begin{table}
    \centering
    \caption{Precision of the extracted pairs\label{tab:precision}}
    \begin{tabular}{c|c|c|c}
        \hline
        method & en-de & en-fr & ro-en\\
        \hline
        ours & \textbf{90.1} & \textbf{95.2} & \textbf{94.5} \\
        FastAlign & 71.3 & 80.0 & 71.8 \\
        \hline
    \end{tabular}
\end{table}

\begin{figure}
    \centering
    \includegraphics[width=0.8\columnwidth]{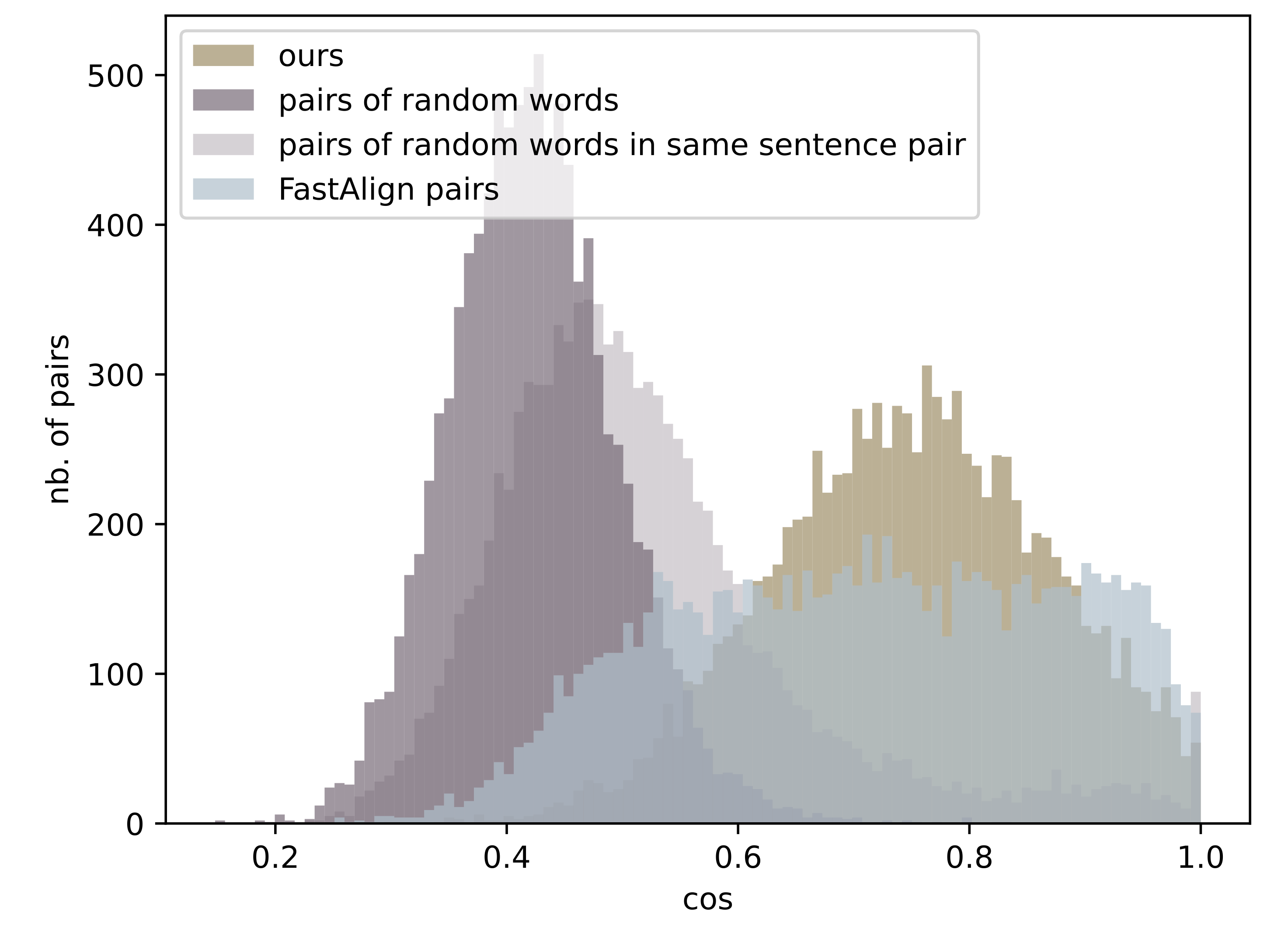}
    \caption{Distribution of similarity between various types of pairs\label{fig:vs_fastalign} (en-de).}
\end{figure}

\cite{zhao-etal-2021-inducing} used FastAlign to compare the similarity of
translated pairs of words and random ones with mBERT. When measuring that
similarity on the last layer and plotting the sampled distribution, they observe
that the pairs obtained with FastAlign give a very broad distribution that
overlaps a lot with random pairs, which leads them to conclude that word-level
representations built by mBERT are not well aligned across languages and it
motivates them to propose a method to re-align representation after pre-training. 

But because they use FastAlign, they are considering many unrelated pairs as
translations. Fig. \ref{fig:vs_fastalign} shows those distributions for the eighth
layer of mBERT. With its right and wrong extracted pairs, the distribution of FastAlign pairs (in blue) overlaps more with the distribution of random pairs (in purple) than the
distribution of pairs extracted with our methods (in yellow). The distribution of random pairs from the same sentence pair (in light purple) is very close to the distribution of random pairs in the whole dataset (in
dark purple), which justifies that any error in the extraction of a pair might increase the overlap between extracted pairs and random pairs.\looseness=-1

This confirms that FastAlign generates too many mistakes to make an accurate
evaluation of the multilingual alignment produced by a model.

\subsection{Multilingual Encoders produce good word-level alignment}

\begin{figure}
    \centering
    \includegraphics[width=0.9\columnwidth]{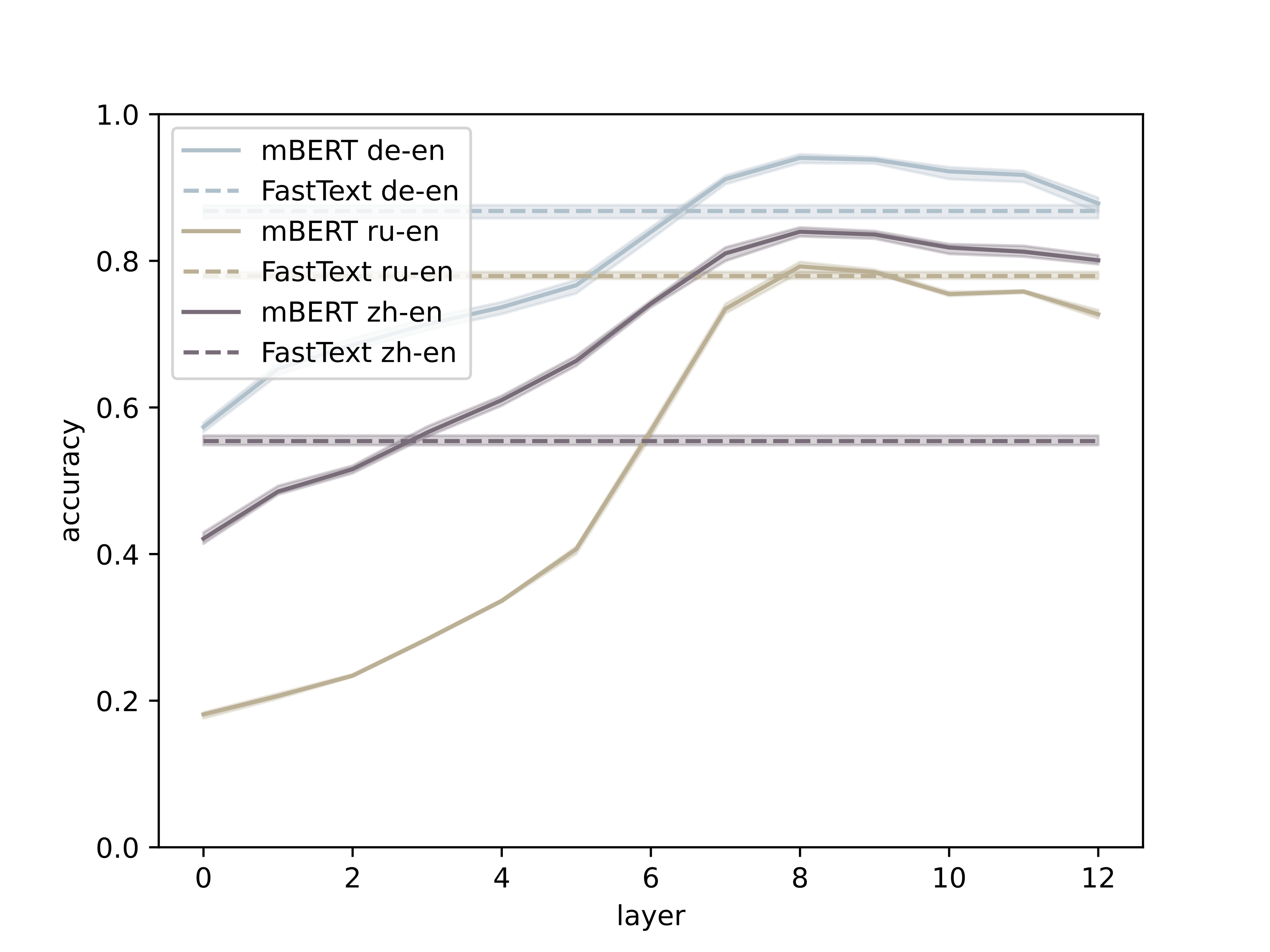}
    \caption{Evolution of $S_{\text{weak}}$ across layers for mBERT compared with FastText aligned.\label{fig:weak_res}}
\end{figure}

Having advocated for our method, we perform the nearest-neighbor search described
in Equation \ref{eq:weak} for evaluating the multilingual alignment. Results for
mBERT and five language pairs are shown in Fig. \ref{fig:weak_res}. We observe
that for a few deep layers, mBERT produces representations that are better
aligned than multilingual word embeddings.

Lower layers give worse alignment. This implies that multilingual
patterns are high-level features. It also goes against the hypothesis
made by several papers
\cite{singh-etal-2019-bert,pires-etal-2019-multilingual,wu-dredze-2019-beto},
that shared vocabulary is what allows models like mBERT to align representations
without having been exposed to parallel texts.

The very last layers also give worse results than layers 8 to 10.
\cite{wu-dredze-2019-beto} have shown that mBERT representations hold
language-specific information at each layer. These language-specific components
might take more importance in the last layers as the pre-training objective is
to predict masked words, a language-specific task. Indeed, the model must learn
not to replace a masked-out word with its translation. 

This type of curve for $S_{\text{weak}}$ is observed for all the multilingual
models we evaluated. Table \ref{tab:weak_res} show retrieval scores for
the first, best and last layer of models described in Section \ref{sec:models}.

\begin{table}
    \centering
    \caption{NN-search results for more models \label{tab:weak_res}}
    \begin{tabular}{l|l|ccc}
\hline
layer & model & de-en & ru-en & zh-en\\
\hline
- & FastText & 86.8 (0.67) & 77.9 (0.39) & 55.4 (0.43)\\
\hline
\multirow{7}{*}{best} & mBERT  & \textbf{94.1} (0.47) & 79.2 (0.64) & \textbf{84.0} (0.40)\\
& XLM-100  & 83.5 (0.52) & 67.4 (0.38) & 27.4 (0.39)\\
& XLM-R Base  & 87.7 (0.41) & 68.7 (0.24) & 63.6 (0.50)\\
& XLM-R Large  & 88.9 (0.30) & 76.8 (0.15) & 72.3 (0.36)\\
& XLM-15$^\mathrm{a,b}$  & 68.5 (0.31) & 28.4 (0.30) & 24.5 (0.56)\\
& AWESOME$^\mathrm{a}$  & 93.4 (0.53) & 76.1 (0.60) & 82.7 (0.38)\\
& mBART$^\mathrm{b,c}$  & 92.1 (0.52) & \textbf{81.2} (0.79) & 74.7 (0.34)\\
\hline
\multirow{7}{*}{last} & mBERT  & 87.9 (0.74) & 72.7 (0.50) & 80.1 (0.43)\\
& XLM-100  & 82.4 (0.60) & 64.6 (0.43) & 25.7 (0.46)\\
& XLM-R Base  & 77.2 (0.81) & 49.3 (0.68) & 51.0 (0.35)\\
& XLM-R Large  & 77.4 (0.49) & 53.5 (0.65) & 54.4 (0.43)\\
& XLM-15$^\mathrm{a,b}$  & 28.5 (0.60) & 4.6 (0.23) & 11.2 (0.39)\\
& AWESOME$^\mathrm{a}$  & 86.5 (0.47) & 67.4 (0.21) & 79.4 (0.44)\\
& mBART$^\mathrm{b,c}$  & 92.1 (0.52) & 81.2 (0.79) & 74.7 (0.34)\\
\hline
\multicolumn{5}{l}{$^\mathrm{a}$ uses parallel data in pre-training $^\mathrm{b}$encodes the language in input}\\
\multicolumn{5}{l}{$^\mathrm{c}$ encoder-decoder model (we only evaluate the encoder)}
    \end{tabular}

\end{table}

The best layer of mBERT gives the best results with respect to all other models.
It might be due to the fact that the next sentence prediction task helps, an
objective on which it is the only pre-trained model. However, it could also
be explained by the fact that mBERT is trained solely on Wikipedia whereas
models like XLM-R are trained on the CommonCrawl corpus which might contain texts
that are less comparable across languages. It is also to be noted that the TLM
objective on parallel texts proposed by the XLM model seems to make the alignment
worse but it is also used in AWESOME.

As the different evaluated models have many differences and obtain somewhat
similar results, one cannot isolate a single parameter that makes the
multilingual alignment better or worse. Nevertheless, it seems that most of
those multilingual models build multilingual representations that are
competitive with word embeddings that have been explicitly aligned. Further
research is needed for identifying the factors that make the quality of such
alignment.

\subsection{Multilingual Encoders produce 'strong' alignment}

\begin{figure}[!t]
    \centering
    \includegraphics[width=0.9\columnwidth]{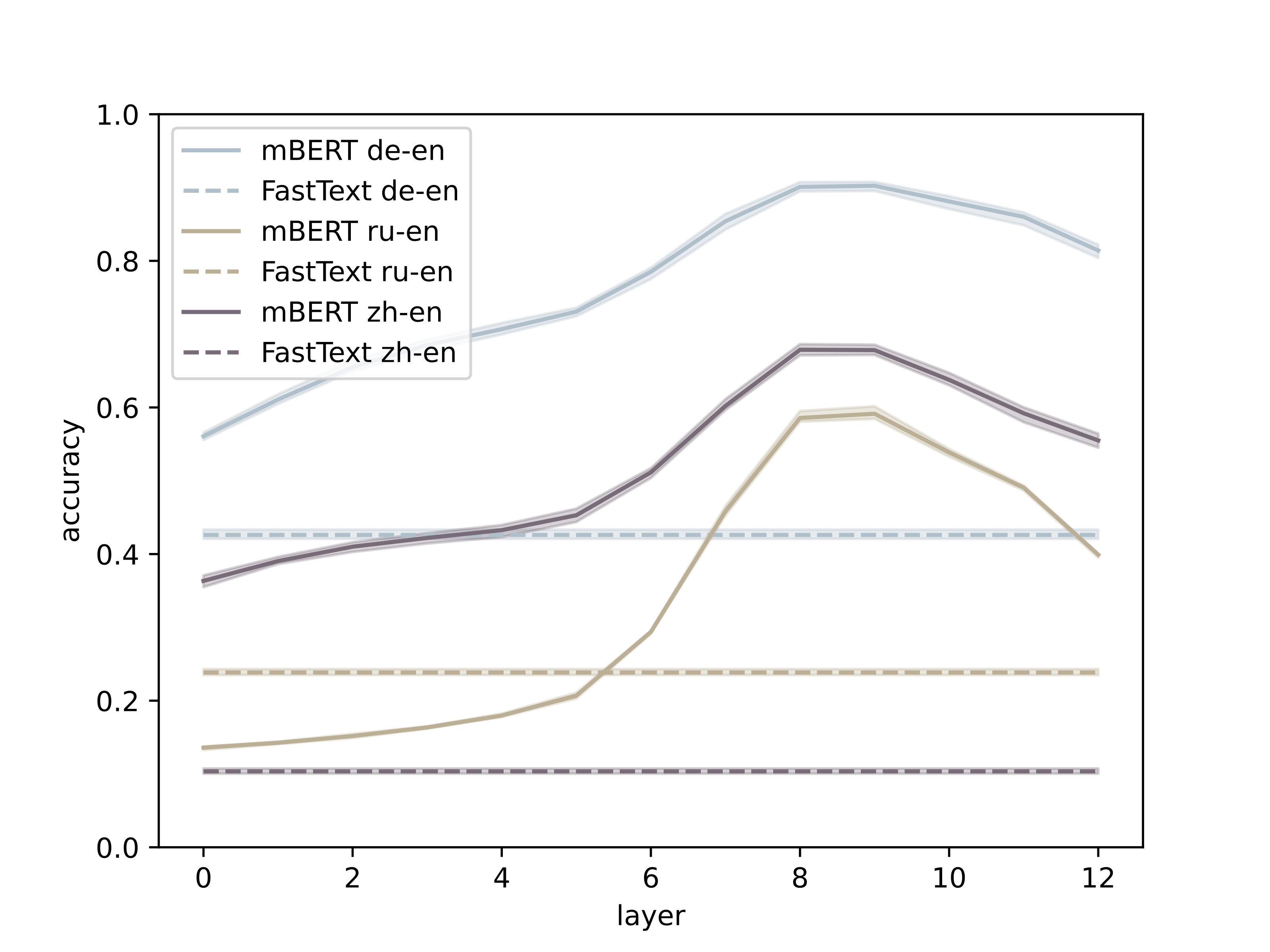}
    \caption{mBERT and FastText aligned representation evaluated with $S_{\text{strong}}$ \label{fig:strong_res}}
\end{figure}

Finally, the same models are evaluated for the strong alignment retrieval
criterion $S_{\text{strong}}$ defined in Equation \ref{eq:strong}. Results for
mBERT are reported on Fig. \ref{fig:strong_res} and results for all models are
reported on Table
\ref{tab:strong_res}.

\begin{table}
    \centering
    \caption{NN-search results for strong alignment \label{tab:strong_res}}
    \begin{tabular}{l|l|ccc}
    
\hline
layer & model & de-en & ru-en & zh-en\\
\hline
- & FastText & 42.6 (0.47) & 23.8 (0.33) & 10.4 (0.23)\\
\hline
\multirow{7}{*}{best} & mBERT  & \textbf{90.3} (0.47) & 59.1 (0.68) & 67.9 (0.49)\\
& XLM-100  & 81.2 (0.49) & 57.9 (0.58) & 22.5 (0.66)\\
& XLM-R Base  & 82.9 (0.46) & 53.5 (0.69) & 49.7 (0.45)\\
& XLM-R Large  & 87.6 (0.40) & \textbf{70.4} (0.61) & 65.1 (0.47)\\
& XLM-15$^\mathrm{a,b}$  & 62.3 (0.63) & 16.7 (0.28) & 21.6 (0.58)\\
& AWESOME$^\mathrm{a}$  & 91.6 (0.82) & 64.6 (0.63) & \textbf{70.9} (0.40)\\
& mBART$^\mathrm{b,c}$  & 88.4 (0.45) & 68.0 (0.82) & 60.0 (0.58)\\
\hline
\multirow{7}{*}{last} & mBERT  & 81.5 (0.63) & 39.9 (0.27) & 55.5 (0.65)\\
& XLM-100  & 72.9 (0.37) & 39.0 (0.58) & 18.6 (0.55)\\
& XLM-R Base  & 73.6 (0.49) & 36.4 (0.65) & 38.6 (0.39)\\
& XLM-R Large  & 72.2 (0.41) & 40.5 (0.50) & 42.6 (0.48)\\
& XLM-15$^\mathrm{a,b}$  & 20.2 (0.79) & 3.7 (0.20) & 8.4 (0.48)\\
& AWESOME$^\mathrm{a}$  & 80.2 (0.77) & 40.5 (0.18) & 56.9 (0.33)\\
& mBART$^\mathrm{b,c}$  & 88.4 (0.45) & 68.0 (0.82) & 60.0 (0.58)\\
\hline
\multicolumn{5}{l}{$^\mathrm{a}$ uses parallel data in pre-training $^\mathrm{b}$encodes the language in input}\\
\multicolumn{5}{l}{$^\mathrm{c}$ encoder-decoder model (we only evaluate the encoder)}
    \end{tabular}
\end{table}

The multilingual alignment of most of the models seems to be robust. Retrieval
accuracy is significantly greater for those multilingual models than for
multilingual word embeddings. There is yet again no way to identify what makes
one Transformer-base model perform better than another. Nevertheless, we have
demonstrated that there is a word-level strong alignment in most multilingual
Transformer-based language models, even for those like mBERT and XLM-R which
have no explicit information about the language in input and haven't been
pre-trained on parallel texts.

\section{Discussion}
From the previous analysis, several questions remain, such as what types of
errors are made and whether this performance is a result of real alignment or
the product of varying densities of sentences of diverse domains. To answer
these questions, we studied a random sample of incorrect predictions on the
WMT19 German-English dataset. As shown in Table \ref{tab:error-analysis},
several error categories are highlighted. The first is due to similar lexical
fields between the target and prediction, which shows that semantically similar examples are clustered together. The second error category shows
that context (e.g. political discourse) can be more important than word
representation, especially for words with low information content, such as
"think". Third, proper nouns are harder to translate, and they are sometimes
associated with other random nouns. Finally, other error patterns are harder to
pinpoint and may be due to density problems.
\begin{table}
    \centering
    \caption{Qualitative analysis of alignment mistakes on examples from WMT19 for the English-German pair \label{tab:error-analysis}}
    \begin{tabular}{l|l|l|p{1.9cm}}
\hline
reference word & target word & model output & error category \\
\hline
angehende & aspiring & genuine & lexical field \\
Anleihe & bond & payout & lexical field \\
Kapitel & chapter & paragraph & lexical field \\
Köpfe & heads & skeletons & lexical field \\
Seen & lakes & fjords & lexical field \\
gestarkt & strengthened & guaranteed & context ("European Union") \\
denke & think & deny & context ("Canada", "European") \\
Bligh & Bligh & Blimp & proper noun \\
Rohrzucker & cane & soldering & \\
Inspektion & inspection & persuasion & \\
standard & default & exclusive & \\
\hline
    \end{tabular}
\end{table}

\begin{figure}[!t]
    \centering
    \includegraphics[width=0.9\columnwidth]{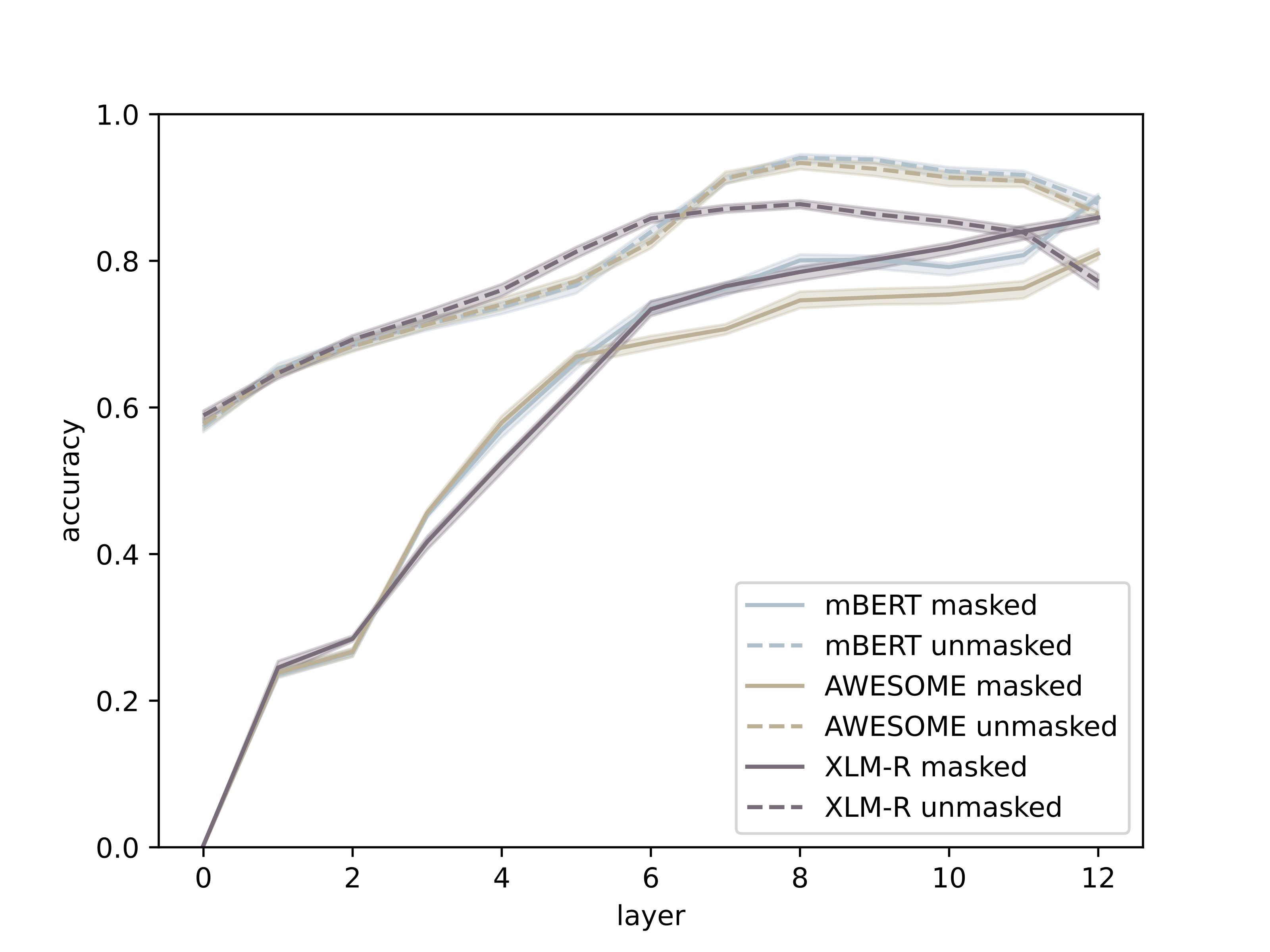}
    \caption{Top-1 accuracy on WMT19 German-English with masked targets \label{fig:masked_res}}
\end{figure}

These considerations suggest how important context is when predicting a
translation. To show this, we studied the performance of models when replacing
the reference and target words with the model mask token, as shown in Figure
\ref{fig:masked_res}. Although performance deteriorates when compared to
performance on the complete sentence, most models perform relatively well
without important information (with a decrease of around 15 points). This
suggests that context is indeed very useful when predicting translations. Since
the tested models are trained with an MLM task, each layer refines the
representation of the mask token, which might explain why performance is
increasing with the depth of representation, contrary to the previous behaviors.
Furthermore, the satisfactory performance on this task suggests that the
extracted representation would be well suited for a cross-lingual retrieval
task. \looseness=-1

\section{Conclusion}

Our results show that Transformer-based models like
mBERT produce contextualized representations of words that are well aligned
across languages, particularly in the deeper layers, despite those models having
only been trained on monolingual objectives. By comparing results for weak and
strong alignment, it was also shown that multilingual Transformers perform far
better on a more challenging evaluation than cross-lingual embeddings which were
built with explicit cross-lingual training signal, like FastText embeddings
aligned with RCSLS but also deeper models trained with TLM objective such as
XLM-15. Finally, our experiments show that averaged
representations are better aligned than representations based on CLS token. 

For future works, we plan to evaluate the extracted representation on retrieval
tasks, as our results show that the studied models seem to be particularly well
suited for such a task. Furthermore, these results are promising for
multilingual Named Entity Linking, which is closely related to a retrieval task,
and could take advantage of data in multiple languages.

\bibliographystyle{IEEEtran}
\bibliography{IEEEabrv,biblio}

\begin{thebibliography}{10}
\providecommand{\url}[1]{#1}
\csname url@samestyle\endcsname
\providecommand{\newblock}{\relax}
\providecommand{\bibinfo}[2]{#2}
\providecommand{\BIBentrySTDinterwordspacing}{\spaceskip=0pt\relax}
\providecommand{\BIBentryALTinterwordstretchfactor}{4}
\providecommand{\BIBentryALTinterwordspacing}{\spaceskip=\fontdimen2\font plus
\BIBentryALTinterwordstretchfactor\fontdimen3\font minus
  \fontdimen4\font\relax}
\providecommand{\BIBforeignlanguage}[2]{{%
\expandafter\ifx\csname l@#1\endcsname\relax
\typeout{** WARNING: IEEEtran.bst: No hyphenation pattern has been}%
\typeout{** loaded for the language `#1'. Using the pattern for}%
\typeout{** the default language instead.}%
\else
\language=\csname l@#1\endcsname
\fi
#2}}
\providecommand{\BIBdecl}{\relax}
\BIBdecl

\bibitem{vaswani-etal-20173}
A.~Vaswani, N.~Shazeer, N.~Parmar, J.~Uszkoreit, L.~Jones, A.~N. Gomez, L.~u.
  Kaiser, and I.~Polosukhin, ``Attention is all you need,'' in \emph{NIPS},
  vol.~30.\hskip 1em plus 0.5em minus 0.4em\relax Curran Associates, Inc.,
  2017.

\bibitem{pires-etal-2019-multilingual}
T.~Pires, E.~Schlinger, and D.~Garrette, ``How multilingual is multilingual
  {BERT}?'' in \emph{Proceedings of ACL}.\hskip 1em plus 0.5em minus
  0.4em\relax ACL, Jul. 2019.

\bibitem{singh-etal-2019-bert}
J.~Singh, B.~McCann, R.~Socher, and C.~Xiong, ``Bert is not an interlingua and
  the bias of tokenization,'' in \emph{EMNLP}, 2019.

\bibitem{dyer-etal-2013-simple}
C.~Dyer, V.~Chahuneau, and N.~A. Smith, ``A simple, fast, and effective
  reparameterization of ibm model 2,'' in \emph{NAACL}, 2013.

\bibitem{zhao-etal-2021-inducing}
W.~Zhao, S.~Eger, J.~Bjerva, and I.~Augenstein, ``Inducing language-agnostic
  multilingual representations,'' \emph{CoRR}, vol. abs/2008.09112, 2020.

\bibitem{devlin-etal-2019-bert}
J.~Devlin, M.-W. Chang, K.~Lee, and K.~Toutanova, ``{BERT}: Pre-training of
  deep bidirectional transformers for language understanding,'' in
  \emph{Proceedings of {NAACL} 2019}, Jun. 2019.

\bibitem{colbert}
O.~Khattab and M.~Zaharia, ``Colbert: Efficient and effective passage search
  via contextualized late interaction over bert,'' 2020.

\bibitem{Bender_2011}
E.~M. Bender, ``On achieving and evaluating language-independence in nlp,''
  \emph{Linguistic Issues in Language Technology}, Oct. 2011.

\bibitem{DBLP:journals/corr/abs-1907-11692}
Y.~Liu, M.~Ott, N.~Goyal, J.~Du, M.~Joshi, D.~Chen, O.~Levy, M.~Lewis,
  L.~Zettlemoyer, and V.~Stoyanov, ``Ro{BERT}a: A robustly optimized {BERT}
  pretraining approach,'' 2020.

\bibitem{Beltagy2019SciBERT}
I.~Beltagy, K.~Lo, and A.~Cohan, ``Scibert: Pretrained language model for
  scientific text,'' in \emph{EMNLP}, 2019.

\bibitem{DBLP:journals/corr/abs-1901-08746}
J.~Lee, W.~Yoon, S.~Kim, D.~Kim, S.~Kim, C.~H. So, and J.~Kang, ``Biobert: a
  pre-trained biomedical language representation model for biomedical text
  mining,'' \emph{Bioinformatics}, vol.~36, no.~4, pp. 1234--1240, 2020.

\bibitem{martin-etal-2020-camembert}
L.~Martin, B.~Muller, P.~J. Ortiz~Su{\'a}rez, Y.~Dupont, L.~Romary,
  {\'E}.~de~la Clergerie, D.~Seddah, and B.~Sagot, ``{C}amem{BERT}: a tasty
  {F}rench language model,'' in \emph{Proceedings of ACL 2020}.\hskip 1em plus
  0.5em minus 0.4em\relax Online: ACL, Jul. 2020.

\bibitem{le-etal-2020-flaubert-unsupervised}
H.~Le, L.~Vial, J.~Frej, V.~Segonne, M.~Coavoux, B.~Lecouteux, A.~Allauzen,
  B.~Crabb{\'e}, L.~Besacier, and D.~Schwab,
  ``\BIBforeignlanguage{English}{{F}lau{BERT}: Unsupervised language model
  pre-training for {F}rench},'' in
  \emph{\BIBforeignlanguage{English}{Proceedings of LREC}}.\hskip 1em plus
  0.5em minus 0.4em\relax ELRA, May 2020.

\bibitem{ronnqvist-etal-2019-multilingual}
S.~R{\"o}nnqvist, J.~Kanerva, T.~Salakoski, and F.~Ginter, ``Is multilingual
  {BERT} fluent in language generation?'' in \emph{Proceedings of the First
  NLPL Workshop on Deep Learning for Natural Language Processing}.\hskip 1em
  plus 0.5em minus 0.4em\relax Turku, Finland: Link{\"o}ping University
  Electronic Press, Sep. 2019.

\bibitem{ralethe-2020-adaptation}
S.~Ralethe, ``\BIBforeignlanguage{English}{Adaptation of deep bidirectional
  transformers for {A}frikaans language},'' in
  \emph{\BIBforeignlanguage{English}{Proceedings of LREC}}.\hskip 1em plus
  0.5em minus 0.4em\relax ELRA, May 2020.

\bibitem{conneau-etal-2020-unsupervised}
A.~Conneau, K.~Khandelwal, N.~Goyal, V.~Chaudhary, G.~Wenzek, F.~Guzm{\'a}n,
  E.~Grave, M.~Ott, L.~Zettlemoyer, and V.~Stoyanov, ``Unsupervised
  cross-lingual representation learning at scale,'' in \emph{Proceedings of
  ACL}.\hskip 1em plus 0.5em minus 0.4em\relax Online: ACL, Jul. 2020.

\bibitem{chomsky_1968}
N.~Chomsky, \emph{Language and Mind}.\hskip 1em plus 0.5em minus 0.4em\relax
  Harcourt Brace, 1968.

\bibitem{greenberg1966language}
J.~H. Greenberg, \emph{Language universals}.\hskip 1em plus 0.5em minus
  0.4em\relax Mouton The Hague, 1966.

\bibitem{wu-dredze-2019-beto}
S.~Wu and M.~Dredze, ``Beto, bentz, becas: The surprising cross-lingual
  effectiveness of {BERT},'' in \emph{Proceedings of EMNLP-IJCNLP}.\hskip 1em
  plus 0.5em minus 0.4em\relax Hong Kong, China: ACL, Nov. 2019.

\bibitem{DBLP:journals/corr/abs-2107-00676}
S.~Doddapaneni, G.~Ramesh, A.~Kunchukuttan, P.~Kumar, and M.~M. Khapra, ``A
  primer on pretrained multilingual language models,'' \emph{CoRR}, vol.
  abs/2107.00676, 2021.

\bibitem{mikolov2013efficient}
T.~Mikolov, K.~Chen, G.~Corrado, and J.~Dean, ``Efficient estimation of word
  representations in vector space,'' \emph{CoRR}, vol. abs/1301.3781, 2013.

\bibitem{DBLP:journals/corr/MikolovLS13}
T.~Mikolov, Q.~V. Le, and I.~Sutskever, ``Exploiting similarities among
  languages for machine translation,'' \emph{CoRR}, vol. abs/1309.4168, 2013.

\bibitem{muse_conneau2017word}
A.~Conneau, G.~Lample, M.~Ranzato, L.~Denoyer, and H.~J{\'e}gou, ``Word
  translation without parallel data,'' \emph{arXiv preprint arXiv:1710.04087},
  2017.

\bibitem{joulin-etal-2018-loss}
A.~Joulin, P.~Bojanowski, T.~Mikolov, H.~J{\'e}gou, and E.~Grave, ``Loss in
  translation: Learning bilingual word mapping with a retrieval criterion,'' in
  \emph{Proceedings of EMNLP}.\hskip 1em plus 0.5em minus 0.4em\relax ACL,
  Oct.-Nov. 2018.

\bibitem{artetxe-etal-2018-robust}
M.~Artetxe, G.~Labaka, and E.~Agirre, ``A robust self-learning method for fully
  unsupervised cross-lingual mappings of word embeddings,'' in
  \emph{Proceedings of ACL}.\hskip 1em plus 0.5em minus 0.4em\relax ACL, Jul.
  2018.

\bibitem{bojanowski2017enriching}
\BIBentryALTinterwordspacing
P.~Bojanowski, E.~Grave, A.~Joulin, and T.~Mikolov, ``Enriching word vectors
  with subword information,'' \emph{CoRR}, vol. abs/1607.04606, 2016. [Online].
  Available: \url{http://arxiv.org/abs/1607.04606}
\BIBentrySTDinterwordspacing

\bibitem{lample-etal-2018-phrase}
G.~Lample, M.~Ott, A.~Conneau, L.~Denoyer, and M.~Ranzato, ``Phrase-based {\&}
  neural unsupervised machine translation,'' \emph{CoRR}, vol. abs/1804.07755,
  2018.

\bibitem{DBLP:journals/corr/abs-1812-10464}
M.~Artetxe and H.~Schwenk, ``Massively multilingual sentence embeddings for
  zero-shot cross-lingual transfer and beyond,'' \emph{CoRR}, vol.
  abs/1812.10464, 2018.

\bibitem{eisenschlos-etal-2019-multifit}
J.~Eisenschlos, S.~Ruder, P.~Czapla, M.~Kadras, S.~Gugger, and J.~Howard,
  ``{M}ulti{F}i{T}: Efficient multi-lingual language model fine-tuning,'' in
  \emph{Proceedings of EMNLP-IJCNLP}.\hskip 1em plus 0.5em minus 0.4em\relax
  ACL, Nov. 2019.

\bibitem{limisiewicz-etal-2020-universal}
T.~Limisiewicz, D.~Mare{\v{c}}ek, and R.~Rosa, ``{U}niversal {D}ependencies
  {A}ccording to {BERT}: {B}oth {M}ore {S}pecific and {M}ore {G}eneral,'' in
  \emph{Findings of EMNLP 2020}.\hskip 1em plus 0.5em minus 0.4em\relax Online:
  ACL, Nov. 2020.

\bibitem{DBLP:journals/corr/abs-2009-12862}
R.~Choenni and E.~Shutova, ``What does it mean to be language-agnostic? probing
  multilingual sentence encoders for typological properties,'' \emph{CoRR},
  vol. abs/2009.12862, 2020.

\bibitem{DBLP:journals/corr/abs-2101-08231}
Z.~Dou and G.~Neubig, ``Word alignment by fine-tuning embeddings on parallel
  corpora,'' \emph{CoRR}, vol. abs/2101.08231, 2021.

\bibitem{roy-etal-2020-lareqa}
U.~Roy, N.~Constant, R.~Al-Rfou, A.~Barua, A.~Phillips, and Y.~Yang,
  ``{LAR}e{QA}: Language-agnostic answer retrieval from a multilingual pool,''
  in \emph{Proceedings of EMNLP}.\hskip 1em plus 0.5em minus 0.4em\relax ACL,
  Nov. 2020.

\bibitem{DBLP:journals/corr/abs-1901-07291}
G.~Lample and A.~Conneau, ``Cross-lingual language model pretraining,''
  \emph{CoRR}, vol. abs/1901.07291, 2019.

\bibitem{DBLP:journals/corr/abs-1809-05053}
A.~Conneau, G.~Lample, R.~Rinott, A.~Williams, S.~R. Bowman, H.~Schwenk, and
  V.~Stoyanov, ``{XNLI:} evaluating cross-lingual sentence representations,''
  \emph{CoRR}, vol. abs/1809.05053, 2018.

\bibitem{DBLP:journals/corr/abs-2001-08210}
Y.~Liu, J.~Gu, N.~Goyal, X.~Li, S.~Edunov, M.~Ghazvininejad, M.~Lewis, and
  L.~Zettlemoyer, ``Multilingual denoising pre-training for neural machine
  translation,'' \emph{CoRR}, vol. abs/2001.08210, 2020.

\bibitem{wmt19translate}
\BIBentryALTinterwordspacing
W.~Foundation. Acl 2019 fourth conference on machine translation (wmt19),
  shared task: Machine translation of news. [Online]. Available:
  \url{http://www.statmt.org/wmt19/translation-task.html}
\BIBentrySTDinterwordspacing

\end{thebibliography}
\end{document}